\documentclass[runningheads]{llncs}
\usepackage{mathtools}
\usepackage{amsmath}
\usepackage{amsfonts}
\usepackage{graphicx}
\begin{document}
	\title{Towards Multi-perspective conformance checking with fuzzy sets}
	\subtitle{Formal and/or Technical paper}
	\author{Sicui Zhang\inst{1,2}\orcidID{0000-0002-5134-7202} 
		\and Laura Genga\inst{2}\thanks{Corresponding author: L.Genga@tue.nl} \orcidID{0000-0001-8746-8826} 
		\and Hui Yan\inst{1,2}\orcidID{0000-0003-0704-7314}
		\and Xudong Lu\inst{1,2}\orcidID{0000-0001-7658-5250}
		\and Huilong Duan\inst{1}
		\and Uzay Kaymak\inst{2,1}\orcidID{0000-0002-4500-9098}
	}
	\authorrunning{S. Zhang et al.}
	
	\institute{School of Biomedical Engineering and Instrumental Science, Zhejiang University, Hangzhou, P.R. China\\
		\and
		School of Industrial Engineering, Eindhoven University of Technology, Eindhoven, The Netherlands\\
	}
	\maketitle              

	\begin{abstract}
		Conformance checking techniques are widely adopted to pinpoint possible discrepancies between process models and the execution of the process in reality. However, state of the art approaches adopt a crisp evaluation of
		deviations, with the result that small violations are considered at the same level of significant ones. This affects the quality of the provided diagnostics, especially when there exists some tolerance with respect to reasonably small violations, and hampers the flexibility of the process. In this work, we propose a novel approach which allows to represent actors' tolerance with respect to violations and to account for severity of deviations when assessing executions compliance. We argue that  besides improving the quality of the provided diagnostics, allowing some tolerance in deviations assessment also enhances the flexibility of conformance checking techniques and, indirectly, paves the way for improving the resilience of the overall process management system.
		\keywords{conformance checking \and fuzzy sets \and data perspective }
	\end{abstract}
	\section{Introduction}
	 \textit{Conformance checking} techniques aim at assessing  to which extent the execution of a process is compliant with respect to a process model representing the expected behavior ~\cite{van2011process}. Since deviating from expected behavior can be costly and/or expose an organization to frauds, conformance checking represents a crucial asset for modern organizations.
State of the art approaches are able both to assess the overall level of compliance of executions and to pinpoint where deviations occurred, thus providing the analyst with valuable diagnostics. 

Nevertheless, nowadays techniques still suffer from some limitations. Among them, in this work we focus on the lack of \textit{flexibility} in compliance analysis.
Processes often involve several alternative execution paths, whose choice can depend on the values of one or more data variables. While this aspect has been traditionally neglected in conformance checking, typically focused on the control flow perspective ~\cite{van2011process,van2012replaying,adriansyah2013memory,adriansyah2012alignment}
, recently few approaches have been proposed to assess process compliance with respect to multiple perspectives \cite{de2013aligning,mannhardt2016balanced}. However, existing techniques consider an activity performed at a given point of an execution either \textit{completely wrong} or \textit{completely correct}. %
Such a crisp distinction is often not suitable in many real-world processes, where decisions on data-guards are often characterized by some level of \textit{uncertainty}, which poses some challenges in drawing exact lines between acceptable/not acceptable values. As a result, in these domains there often exists some tolerance to deviations. For example, let us assume that in a medical process there is a guideline stating that in between two procedures there must be an interval of at most five hours. Adopting a crisp evaluation, 4 hours 59 minutes would be considered fully compliant, while 5 hours and 1 minute would be fully not compliant, which is intuitively unreasonable. Such an approach can lead to generating misleading diagnostics, where executions marked as deviating actually correspond to acceptable behaviors. 
Furthermore, the magnitude of the deviations is not considered; small and large deviations are considered at the same level of compliance, which can easily be misleading. It is worth noting that this approach can also hamper the overall process resilience, making it very sensible even to small exceptions/disruptions. For instance, if process executions are monitored real-time, every small deviations can lead to raise some alarms and/or to stop the execution. 

To deal with these challenges, in the present work we perform an exploratory study on the use of \textit{fuzzy sets}\cite{cheng1992fuzzy} in conformance checking. Fuzzy sets have been proven to be a valuable asset to represents human decisions making process, since they allow to formalize the uncertainty often related to these processes.  In particular, elaborating upon fuzzy sets concepts, we propose a new multi-perspective conformance checking technique that accounts for the degree of deviations. Taking into account the severity of the occurred deviations allows to a) improving the quality of the provided diagnostics, generating a more accurate assessment of the deviations, and b) enhancing the flexibility of compliance checking mechanisms, thus paving the way to improve the robustness of the process management system with respect to unforeseen exceptions, that is a necessary step towards the development of \textit{resilient} systems~\cite{muller2013resilience}. 
As a proof-of-concept, we tested the approach over a synthetic dataset. 

The rest of this work is organized as follows. Section \ref{sec:related} discusses related work; Section \ref{sec:preliminaries} introduces basic concepts used throughout the paper; Section \ref{sec:motivation} introduces a running example to discuss the motivation of this work; Section \ref{sec:methodology} illustrates the approach; Section \ref{sec:exp} discusses results obtained by a set of synthetic experiments; finally, Section \ref{sec:conclusion} draws some conclusions and future work.
	\section{Related work}
	\label{sec:related}
	During last decades, several conformance checking techniques have been proposed.
Some approaches \cite{borrego2014conformance,caron2013comprehensive,taghiabadi2014compliance} propose to check whether event traces satisfy a set of compliance rules, typically represented using declarative modeling.
Rozinat and van der Aalst \cite{rozinat2008conformance} propose a token-based technique to replay event traces over a process model to detect deviations.Although this technique can deal with infinite behavior, it has been shown that token-based techniques can provide misleading diagnostics \cite{adriansyah2010towards}.
Recently, alignments have been proposed as a robust approach to conformance checking \cite{van2012replaying}.
Alignments are able to pinpoint deviations causing nonconformity based on a given cost function. While most of alignment-based approaches use the standard distance cost function as defined by \cite{van2012replaying}, some variants have been proposed to enhance the quality of the provided diagnostics. For example,  the work of
Alizadeh et al.\ \cite{alizadeh2014history} proposes an approach to compute the cost function  by analyzing historical logging data, with the aim of  obtaining probable explanations of nonconformity.
Besides the control flow, there are also other perspectives like data, or resources, that are often crucial for compliance checking analysis. Few approaches in literature have investigated how to include these perspectives in the analysis. \cite{alizadeh2015constructing} extends the approach in \cite{alizadeh2014history}, to enhance the accuracy of the probable non-conformity explanations by taking into account data describing the contexts in which the activities occurred in previous process executions.
Some approaches proposed to compute the control-flow firstm then assessing process executions compliance with respect to the data perspective, e.g., ~\cite{de2013aligning}, \cite{hoffmansapproach}. These methods assume that the control flow is more important than other perspectives for an optimal alignment, with the result that some important deviations can be missed.
~\cite{mannhardt2016balanced} introduces a cost function able to account for all kind of deviations at the same time, thus obtaining well-rounded diagnostics considering all the desired perspectives.
The approaches mentioned so far  assume a crisp evaluation of deviations, according to which a deviation is completely wrong or completely correct. In this work, we aim at considering the severity of the detected deviations by using fuzzy sets notions. Several studies in literature have proven that fuzzy sets can be successfully employed  to represent humans' decision making processes; among them, we can mention, for example \cite{bosma2005} to study a fuzzy approach to modelling Vietnames farmers' decision process in adopting adopting integrated farming systems. However, to the best of our knowledge, no previous work has investigated the use of fuzzy sets concept for conformance checking.
	\section{Motivating Example}
	\label{sec:motivation}
\begin{figure}[h]
\centering
\includegraphics[width=1.05\linewidth]{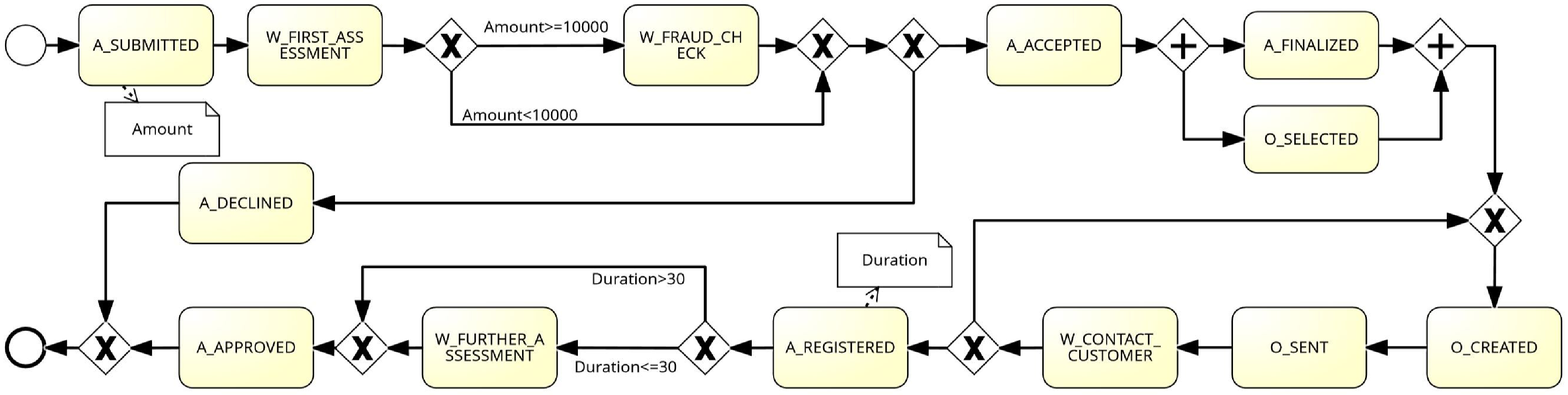}
\caption{The Load Management Model.}
\label{fig1}
\end{figure}

	Consider, as a running example, a loan management process   derived from previous work on the event log of a financial institute made available for the BPI2012 challenge \cite{adriansyah2012mining,genga2018discovering}. 
Fig.~\ref{fig1} shows the process in BPMN notation.
The process starts with the submission of an application. Then, the application passes through a first assessment, aimed to verify whether the applicant meets the requirements.
If the requested amount is greater than 10000 euros, the application also goes through a more accurate analysis to detect possible frauds. If the application is not eligible, the process ends; otherwise, the application is accepted. An offer to be sent to the customer is selected and the details of the application are finalized. After the offer has been created and sent to the customer, the latter is contacted to discuss the offer with him/her, possibly adjusting according to her preferences. 
At the end of the negotiation, the agreed application is registered on the system. At this point, further checks can be performed on the application, if the overall duration is still below 30 days, before approving it.

Let us consider the following example traces\footnote{We use the notation $(act,\{att_1=v_1,\ldots, att_n=v_n\})$ to denote the occurrence of activity $act$ in which variables $att_1 \ldots att_n$ are assigned to corresponding values $v_1, \ldots v_n$}: $\sigma_1=\langle (A\_S,\{Amount$ $=9950\}),$ $(W\_FIRST\_A,\bot),$ $(W\_F\_C,\bot),$ $(A\_A,\bot),$ $(A\_F,\bot),$ $(O\_S,\bot),$ $(O\_C,\bot),\\$ $(O\_S,\bot),$ $(W\_C,\bot),$ $(A\_R,\{Duration=50\}),$ $(A\_{AP},\bot),$  $\rangle$ ; \\$\sigma_2=\langle (A\_S,\{Amount$ $=2000\}),$ $(W\_First\_A,\bot),$ $(W\_F\_C,\bot)$ $(A\_A,\bot),$ $(A\_F,\bot),\\$ $(O\_S,\bot),$ $(O\_C,\bot),$ $(O\_S,\bot),$ $(W\_C,\bot),$ $(A\_R,\{Duration=60\}),$ $(A\_{AP},\bot),$ $\rangle$. 
Both these executions violate the guard on the $Amount$ value; indeed, the activity $W\_F\_C$ should have been skipped, being the requested loan amount lower than 10000.
It is worth noting, however, that there is a significant difference in terms of their \textit{magnitude}. Indeed, while in the first execution the threshold was not reached only by few dozens of euros, the second violation is several thousands of euros below the limit.
Since state-of-the art conformance checking techniques adopt a \textit{crisp} logic, where the value of a data variable can be marked only either as correct or wrong, this difference between $\sigma_1$ and $\sigma_2$ remains undetected.

We argue that taking into account the severity of the violations when assessing execution compliance allows to obtain more accurate diagnostics, especially in contexts where there exists some uncertainty related to the guards definition. Indeed, in these cases guards often represent more guidelines, rather than strict, sharp rules, and there might be some tolerance with respect to violations. In our example, $\sigma_1$ could model an execution considered suspicious for some reasons, making a a fraud check worthy, since the amount is only slightly less than 10000. On the other hand, the violation in $\sigma_2$ desreves some attention, since the amount is so far from the threshold that the additional costs needed for the fraud check are probably not justified.

Differentiating among different levels of violations also impacts the interpretation of the deviations. Often, multiple interpretations are returned by conformance checking techniques. For example, in our case possible interpretations can be 1) the activity $W\_F\_C$ should have been skipped, or 2) the execution of the activity is correct but it occurred with unexpected value of the variable $Amount$. Differentiating between the severity of the deviations would make the second interpretation the preferred one when the deviation is limited, like in $\sigma_1$, thus providing more guidance to the analyst during process diagnostics. 

	\section{Preliminaries}
	\label{sec:preliminaries}
	This section introduces a set of definitions and concepts that will be used through the paper. First, we recall important \textit{conformance checking} notions; secondly, we introduce basic elements of \textit{fuzzy sets theory}.
\subsection{Conformance Checking: Aligning Event Logs and Models}
Conformance checking techniques detect discrepancies between a process model describing the expected process behavior and the real process execution.

The expected process behavior is typically represented as a \textit{process model}. Since the present work is not constrained to the use of a specific modeling notation, here we refer to the notation used in \cite{van2012replaying}, enriched with data-related notions explained in \cite{de2013aligning}. 
\begin{definition}[Process model]
A process model $M=(P,P_I,P_F,A_M, V, U,\\ T, G, W, Values)$ is a transition system defined over a set of activities $A_M$ and a set of variables $V$, with states $P$, initial states $P_I \subseteq P$, final states $P_F \subseteq P$ and transitions $T \subseteq P\times (A_M \times 2^V) \times P$. The function $U$ defines the admissible data values, i.e.,  $U(V_i)$ represents the domain of $V_i$ for each $V_i \in V$; the function $G : A_M \rightarrow Formulas(V \cup \{V_i' \mid V_i\in V\})$ is a \textit{guard} function, that associates an activity to a guard, i.e., a boolean formula expressing a condition on the values of the data variables; $W: A_M \rightarrow 2^V$ is a write function, that associates an activity with the set of variables which are written/updated by the activity; finally, $Values : P\rightarrow \{V_i=v_i, i=1..|V| \mid v_i \in U(V_i) \cup \{\bot\} \}$ is a function that associates each state with the corresponding pairs variable=value.
\end{definition}
When a variable $V_i \in V$ appears in a guard $G(A_M)$, it refers to the value just before the occurrence of $A_M$; however, if $V_i \in W(A_M)$, it can also appear as $V_i'$, and refers to the value after the occurrence.
The firing of an activity $s=(a,w) \in A_M \times (V \not\rightarrow U)$ in a state $p'$ is \textit{valid} if: 1) $a$ is enabled in $p'$; 2) $a$ writes all and only the  variables in $W(a)$; 3) $G(a)$ is $true$ when evaluate over $Values(p')$. To access the components of $s$ we introduce the following notation: $vars(s)=w$, $act(s)=a$. Function $vars$ is also overloaded such that $vars(s,V_i)=w(V_i)$ if $V_i \in dom(vars(s))$ and $vars(s,V_i)=\bot$ if $V_i \not\in dom(vars(s))$.
The set of valid process traces of a process model $M$ is denoted with $\rho(M)$ and consists of all the valid firing sequences $\sigma \in (A_M \times (V \not\rightarrow U))^*$ that, from an initial state $P_I$ lead to a final state $P_F$.

Process executions are often recorded by means of an information system in so-called \textit{event logs}. In particular, an event log consists of \textit{traces}, each collecting the sequence of events recorded during the same process execution. Formally, let $S_N$ be the set of (valid and invalid) firing of activities of a process model $M$; an \textbf{event log} is a multiset of traces $\mathbb{L} \in \mathbb{B}(S_N^*)$.
Given an event log $L$, conformance checking builds an $alignment$ between $L$ and $M$, whose goal consists in relating activities occurred in the event log to the activities in the model and vice versa. To this end, we need to map \lq\lq moves" occurring in the event log to possible \lq\lq moves" in the model. However, since the executions may deviate from the model and/or not all activities may have been modeled or recorded~\cite{van2012replaying}, we might have log/model moves which cannot be mimicked by model/log moves respectively. These situations are modeled by a \lq\lq no move" symbol \lq\lq $\gg$ ". 
For convenience, we introduce the set $S_N^\gg=S_N \cup \{\gg\}$.
Formally, we set $s_L$ to be a transition of the events in the log, $s_M$ to be a transition of the activities in the model. A move is represented by a pair $(s_L,s_M) \in S_N^\gg \times S_N^\gg$ such that:
\begin{itemize}
    \item $(s_L,s_M)$ is a \textit{move in log} if $s_L \in S_N$ and $s_M\;=\;\gg$
\item $(s_L,s_M)$ is a \textit{move in model} if $s_M \in S_N$ and $s_L\;=\;\gg$
\item  $(s_L,s_M)$ is a \textit{move in both without incorrect write operations} if $s_L \in S_N$, $s_M \in S_N$ and $act(s_L)=act(s_M)$ and $\forall V_i \in V (vars(s_L,V_i)= vars(s_M,V_i)))$
\item $(s_L,s_M)$ is a \textit{move in both with incorrect write operations} if $s_L \in S_N$, $s_M \in S_N$ and $act(s_L)=act(s_M)$ and $\exists V_i \in V \mid vars(s_L,V_i) \neq vars(s_M,V_i))$ 
\end{itemize}

Let $ A_{LM}=\{(s_L,s_M) \in S_N^\gg \times S_N^\gg \mid s_L \in S_N \lor s_M \in S_N \} $ be the set of all legal moves. The \textit{alignment} between two process executions $ \sigma_L, \sigma_M  \in  S_N^* $ is $ \gamma \in A_{LM}^* $ such that the projection of the first element (ignoring $ \gg $) yields $ \sigma_L $, and the projection on the second element (ignoring $ \gg $) yields $ \sigma_M $.
\begin{example}
\label{ex:ex_alignment}
Let us consider the simple model represented in Fig.~\ref{exampleModel} and the trace $\sigma_L=\langle (a,\{V_1=35\}), (b,\bot), (c,\bot)  \rangle$.
Table~\ref{t1} shows two possible alignments $\gamma_1$ and $\gamma_2$ for $\sigma_L$. For $\gamma_1$, the pair $(b,b)$ is a move in both with incorrect data, since the value of $V_1$ is not allowed when activity $b$ is executed; while in $\gamma_2$ the move $(\ b,\bot)$  is matched with a $\gg$, i.e., it is a move on log.
\begin{figure}[h]
\centering
\includegraphics[scale=0.3]{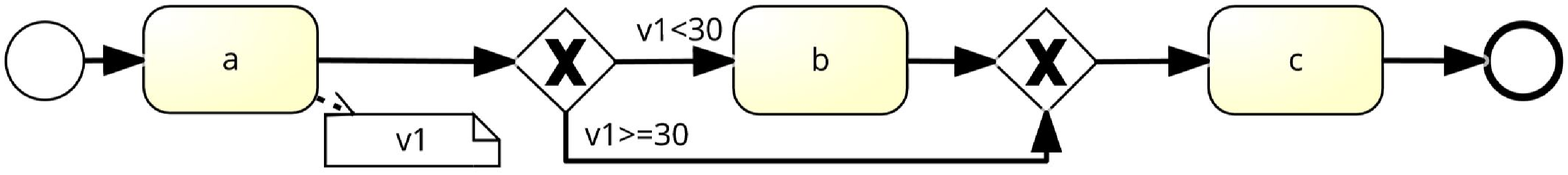}
\caption{A Simple Example Model}
\label{exampleModel}
\end{figure}
\begin{table}
\centering
\caption{Two possible alignments between $\sigma_M$ and $\sigma_L$}\label{t1}
       \begin{tabular}{|l|l|l|l|}
	\hline
	\multicolumn{2}{|l}{Alignment $ \gamma_1$ } &  \multicolumn{2}{|l|}{Alignment $ \gamma_2$ }\\
\hline
Log &  Model &  Log &  Model\\
$(a,\{V_1=35\})$ &  $(a, \{V_1\})$ &  $(a, \{V_1=35\})$ &  $(a, \{V_1 \})$\\
$(b, \bot)$ &  $(b\ \bot)$&  $(b, \bot)$ & $\gg$ \\
$(c, \bot)$ & $(c \ \bot)$ &  $(c, \bot)$&  $(c \ \bot)$  \\
\hline
	\end{tabular}
\end{table}
\end{example}

As shown in Example \ref{ex:ex_alignment}, there can be multiple possible alignments for a given log trace and process model. Our goal is to find the \textit{optimal alignment}, i.e., a complete alignment as close as possible to a proper execution of the model. To this end, the severity of deviations is assessed by means of a \textit{cost function}:
\begin{definition}[Cost function, Optimal Alignment]
Let $\sigma_L$, $\sigma_M$ be a log trace and a model trace, respectively. Given the set of all legal moves $A_N$, a \textit{cost function} $k$ assigns a non-negative cost to each legal move: $A_N \rightarrow \mathbb{R}_0^+$. The \textit{cost of an alignment $\gamma$} between $\sigma_L$ and $\sigma_M$ is computed as the sum of the cost of all the related moves: $\mathit{K(\gamma)}=\sum_{(S_L,S_M) \in \gamma}k(S_L,S_M)$. An \textbf{optimal} alignment of a log trace and a process trace is one of the alignments with the lowest cost according to the provided cost function.
\end{definition}
\subsection{Basic Fuzzy Sets Concepts}
Classic sets theory defines crisp, dichotomous functions to determine membership of an object to a given set. For instance, a set N of real numbers smaller than 5 can be expressed as $N = \{n \in \mathbb{R}  |n < 5\}$. In this setting, an object either belongs to $N$ or it does not.
Although crisp sets have proven to be useful in various applications, there are some drawbacks in their use. In particular, human thoughts and decisions are often characterized by some degree of uncertainty and flexibility, which are hard to represent in a crisp setting \cite{jang1997neuro}.

\textit{Fuzzy sets theory} aims at providing a meaningful representation of measurement uncertainties, together with a meaningful representation of vague concepts expressed in natural language and close to human thinking ~\cite{klir1995fuzzy}. Formally, a \textit{fuzzy set} is defined as follows:

\begin{definition}[Fuzzy Set]
Let $N$ be a collection of objects. A \textit{fuzzy set} $F$ over $N$ is defined as a set of ordered pairs $F=\{n, \mu_F(n) \mid n \in N)\}$. $\mu_F(n)$ is called the membership function (MF) for the fuzzy set F, and it's defined as $\mu_F: N \to [0,1]$. The set of all points $n$ in $N$ such that $\mu_F(n) > 0$ is called the \textbf{support} of the fuzzy set, while the set of all points in $N$ in which $\mu_F(n)=1$ is called \textbf{core}.
\end{definition}

It is straightforward to see that fuzzy sets are extensions of classical sets, with the characteristic function allowing to any value between 0 and 1. 
In literature several standard functions have been defined for practical applications (see, e.g., \cite{klir1995fuzzy} for an overview of commonly used functions).

	\section{Methodology}
	\label{sec:methodology}
	The goal of this work is introducing a compliance checking approach tailored to take into account the severity of the deviations, in order to introduce some degree of flexibility when assessing compliance of process executions and to generate diagnostics more accurate and possible closer to human interpretation. To this end, we investigate the use of \textit{fuzzy sets theory}. %
In particular, we propose to use fuzzy membership functions to model the cost of moves involving data; then, we employ off-shelf techniques based on the use of A* algorithm to  build the optimal alignment.
The approach is detailed in the following subsections. %
\subsection{Fuzzy cost function}
\label{subsec:cf}
The computation of an optimal alignment relies on the definition of a proper cost function for the possible kind of moves (see Section \ref{sec:preliminaries}). Most of state-of-the art approaches adopt ( variants of) the standard distance function defined in \cite{van2012replaying}, which sets a cost of 1 for every move on log/model (excluding invisible transitions), and a cost of 0 for synchronous moves. Furthermore, the analyst can use \textit{weights} to differentiate between different kind of moves. 
%

The standard distance function is defined only accounting for the control-flow perspective. However, in this work we are interested in the data-perspective as well. In this regards, a cost function explicitly accounting for the data perspective has been introduced by \cite{mannhardt2016balanced} and it is defined as follows.
\begin{definition}[Data-aware cost function]
\label{def:data_cost_function}
Let $(S_L,S_M)$ be a move between a log trace and a model execution, and let, with a slight abuse of notation, $W(S_M)$ to represent write operations related to the activity related to  $S_M$.
The cost $k(S_L,S_M)$ is defined as:

\begin{equation}
\label{k(S_L,S_M)_crisp}
k(S_L,S_M)=
\begin{cases}
1 & \; \text{if} \: (S_L, S_M) \: \text{is a move in log}\\
1+ |W(S_M)| & \; \text{if} \: (S_L, S_M) \: \text{is a move in model}\\
|\{V_i \in W(S_M) : & \; \text{if} \: (S_L, S_M) \: \text{is a move in both with}  
\\ var(S_L,V_i) \neq var(S_M,V_i)\}| & \;  \text{incorrect write operations}\\
0 & \text{otherwise}
\end{cases}
\end{equation}
\end{definition}
In the previous definition, data costs are computed as a)number of data variables not written/updated because the corresponding activity was skipped , b) number of data variables in a move whose values are not allowed according to the process model. 
The previous function considers every move either as \textit{completely wrong} or \textit{completely correct}; namely, it is a dichotomous function. To differentiate between different magnitude of deviations, in this work we propose to use fuzzy membership functions as cost functions for the alignment moves. Note that here we focus on data moves. Indeed, when considering other perspectives the meaning of the severity of the deviation is not that straightforward. For example, when considering  control-flow deviations, usually an activity is either executed or skipped. Nevertheless, fuzzy costs can be defined also for other process perspectives, for instance, %
to differentiate between skip of activities under different conditions. We plan to explore these directions in future work.

Following the above discussion, we define our \textit{fuzzy cost function} as follows:

\begin{definition}[Data-aware fuzzy cost function]
\label{def:fuzzycost}
Let $(S_L,S_M)$ be a move between a process trace and a model execution, and let $MF(var(S_L,V_i))$ be a fuzzy membership function returning the degree of deviation of a data variable in a move with incorrect data. The cost $k(S_L,S_M)$ is defined as:

\begin{equation}
\label{k(S_L,S_M)}
k(S_L,S_M)=
\begin{cases}
1 & \; \text{if} \: (S_L, S_M) \: \text{is a move in log}\\
1+ |W(S_M)| & \; \text{if} \: (S_L, S_M) \: \text{is a move in model}\\
\sum_{\forall V_i \in V} MF(var(S_L,V_i))  & \; \text{if} \: (S_L, S_M) \: \text{is a move in both with} \\ 
& \text{incorrect write operations}\\
0 & \text{otherwise}
\end{cases}
\end{equation}
\end{definition}

To define the fuzzy cost function in (\ref{k(S_L,S_M)}), we first need to determine over which data constraints we want to define a $MF$ \footnote{Note that multiple $MF$ functions can be defined for the same data variable, if it is used in multiple guards.}. Then, for each of them first we need to define a tolerance interval; in turn, this implies to define a) an interval for the core of the function, and b) an interval for the support of the function, (see Section \ref{sec:preliminaries}). This choice corresponds to determine, for a given data constraint, which values should be considered equivalent and which ones not optimal but still acceptable. Once the interval is chosen, we need to select a suitable membership function.
In literature, several different $MF$ have been defined (see, e.g., \cite{klir1995fuzzy} for an overview), with different level of complexity and different interpretations. 
It is straightforward to see that determining the best $MF$ to explicit the experts' knowledge is not a trivial task. For the sake of space, an extended discussion over the $MF$ modeling is out of the scope of this paper, and left for future work. Nevertheless, we would like to point out that this is a well-studied issue in literature, for which guidelines and methodologies have been drawn like, e.g., the one presented by \cite{Cornelissen2003}. The approach can be used in combination of any of these methodologies, since it does not depend on the specific $MF$ chosen.

\begin{example}
\label{ex:mf_function}
Let us consider again the alignment $\gamma_1$ in Table~\ref{t1} and the model in Fig.~\ref{exampleModel}. According to a crisp cost function, the cost for the second move would be 1, since variable $V_1$ does not fulfill the corresponding guard. Now, let us assume to interview an expert of this process, who tells us that values of $V_1$ up to 40 are still acceptable, even though not optimal. Let us represent this knowledge using a so-called R-function as $MF$, that are commonly used for their simplicity when no further information is available, defined as follows. 
\begin{equation*}\label{MF(vars(b,v_1))}
MF(vars(b,V_1))=
\begin{cases}
0 & \; \text{if} \: vars(b,V_1) \geq v_{ub}\\
\frac{v_{ub} - vars(b,v_1)}{v_{ub} - v_o} & \text{if} \; v_o < vars(b,V_1) < v_{ub} \\ 
1 & \text{if} \; vars(b,v_1) < v_o\\
\end{cases}
\end{equation*}
where $v_{ub}$ represents the upper bound the analyst is willing to accept, while $v_o$ represents the ideal value represented by the constraint. With $vars(b,V_1)=35$, $v_o=30$ and $v_{ub}=40$, we would obtain a move cost equal to 0.5. 
\end{example}

\subsection{Alignment building: using A* to find the optimal alignment}
\label{subsec:ab}

The problem of finding an optimal alignment is usually formulated as a search problem in a directed graph
\cite{dechter1985generalized}. 
Let $Z=(Z_V, Z_E)$ be a directed graph with edges weighted according to some cost structure. The A* algorithm finds the path with the lowest cost from a given source node $v_0 \in Z_v$ to a node of a given goals set $Z_G \subseteq Z_V$. The coast for each node is determined by an evaluation function 
$f(v)=g(v)+h(v)$,    
where:
\begin{itemize}
    \item $g: Z_V \rightarrow \mathbb{R}^+$ gives the smallest path cost from $v_0$ to $v$;
    \item $h: Z_V \rightarrow \mathbb{R}_0^+$ gives an estimate of the smallest path cost from $v$ to any of the target nodes.
\end{itemize}
If $h$ is \textit{admissible},i.e. t underestimates the
real distance of a path to any target node $v_g$, A* finds a path that is guaranteed to have the overall lowest cost.

The algorithm works iteratively: at
each step, the node v with lowest cost is taken from a priority queue. If v
belongs to the target set, the algorithm ends returning node v. Otherwise, v is
expanded: every successor $v_0$ is added to priority queue with a cost $f(v_0)$.

Given a log trace and a process model, to employ A* to determine an optimal alignment we associate every node of the search space with a prefix of some complete alignments. 
The source node is
an empty alignment $\gamma_0 = \langle \rangle$, while the set of target nodes includes every complete
alignment of $\sigma_L$ and $M$. For every  pair of nodes $(\gamma_1, \gamma_2)$, $\gamma_2$ is obtained by adding one move to $\gamma_1$.

The cost associated with a path leading to a graph node $\gamma$ is then defined as
$g(\gamma)= K(\gamma) + \epsilon|\gamma|$, 
where $K(\gamma)=\sum_{(s_L,s_M \in \gamma}k(s_L, s_M)$, with $k(s_L, s_M)$ defined as in (\ref{k(S_L,S_M)}); $|\gamma|$ is the number of moves in the alignment; and $\epsilon$ is a negligible cost, added to guarantee termination. Note that tthe cost $g$ has to be strictly increasing. While a formal proof is not possible for the sake of space, it is however straight to see that $g$ is obtained in our approach by the sum of all non negative elements; therefore, while moving from an alignment prefix to a longer one, the cost can never decrease.
For the definition of the heuristic cost function $h(v)$ different strategies can be adopted. Informally, the idea is computing, from a given alignment, the minimum number of moves (i.e., the minimum cost) that would lead to a complete alignment. Different strategies have been defined in literature, e.g., the one in \cite{van2012replaying}, which exploits Petri-net marking equations, or the one in \cite{yan2017aligning}, which generates possible states space of a BPMN model.

	\section{Implementation and Experiments}
	\label{sec:exp}
	This section describes a set of experiments we performed to obtain a proof-of-concept of the approach. To this end, we compared the diagnostics returned by a crisp conformance checking approach with the outcome obtained by our proposal. In order to get meaningful insights on the behavior we can reasonably expect by applying the approach in the real world,  we employ a realistic synthetic event log, introduced in a former paper~\cite{genga2019predicting}, obtained starting from one real-life logs, i.e., the event log of the BPI2012 challenge \footnote{\url{https://www.win.tue.nl/bpi/doku.php?id=2012:challenge}}. We evaluated the compliance of this log against a simplified version of the process model in~\cite{genga2019predicting}, to which we added few data constraints (see Fig.\ref{fig1}). The approach has been implemented as an extension to the tool developed by~\cite{yan2017aligning}, designed to deal with BPMN models. 
In the following we describe the experimental setup and the obtained results.
\paragraph{Settings:} The log in~\cite{genga2019predicting} consists of 5000 traces, where a predefined set of deviations was injected. The values for the variables "$Amount$" were collected the from the BPI2012 log, while for calculating "$Duration$" a random time window ranging from 4 to 100 hours has been put in between each pair of subsequent activities, and the overall duration was then increased of by 31 days for some traces. For more details on the log construction, please check ~\cite{genga2019predicting}.

Our process model involves two constraints for the data perspective, i.e.,  $Amount >= 10000$ to execute the activity $W\_F\_C$, and $Duration <= 30$ to execute  the activity $W\_FURTHER\_A$.
For the crisp conformance checking approach, we use the cost function provided by 
(\ref{k(S_L,S_M)_crisp}); 
while for the fuzzy approach, the cost function in 
(\ref{k(S_L,S_M)}).

Here we assume that $Amount \in (3050,10000)$ and $Duration \in (30,70)$ represent a tolerable violation range for the variables. Since we cannot refer to experts' knowledge here, we derived these values from simple descriptive statistics. In particular, we draw the distributions of the values for each variable, considering values falling within the third quartile as acceptable. The underlying logic is that values which tend to occur repeatedly are likely to indicate acceptable situations. Regarding the shape of the membership function, here we apply the following R function explained in Example~\ref{ex:mf_function}, reported below. $Amount$ and $Duration$ are abbreviated to $A$ and $D$.

\hbox to \hsize{\vbox{\hsize=0.5\hsize 
\begin{equation}\nonumber\label{S1Amount}
MF_1(A)=
      \begin{cases}
      0 & \text{, if $\ A \geq 10000 $}\\
      1 & \text{, if $\ A \leq 3050 $}\\
       \frac{10000-A}{6950}& \text{, if $\ 3050 < A < 10000 $;}\\
    \end{cases} 
\end{equation}}
\vbox{\hsize=0.4\hsize \begin{equation}\nonumber\label{S2Dura}
MF_2(D)=
      \begin{cases}
      0 & \text{, if $\ D \leq 30 $}\\
      1 & \text{, if $\ D \geq 70 $}\\
      \frac{D - 30}{40}& \text{, if $\ 30 < D < 70$}\\
    \end{cases} 
\end{equation}}}
\paragraph{Results:} We compare the diagnostics obtained by the crisp approach and by our approach in terms of a)kind of moves regarding the activities ruled by the guard, and b)distribution of fitness values, computed according to the definition in \cite{de2013aligning}. 
Table~\ref{t2_alignment} shows differences in terms of number and kind of moves detected for the activities $W\_F\_C$ and $W\_FURTHER\_A$ within the crisp/fuzzy alignments respectively, considering also the possible existence of multiple optimal alignments. Namely, when the same move got different interpretations in different alignments, we count the move as both move in log and move in data. Note, however, that multiple optimal alignments with the same interpretation for the move count one. It is worth noting that while we obtained the same number of move-in-log, move-in-data for the crisp approach, these values change when considering the fuzzy approach, for which move in log are in general less. Indeed, the alignments obtained by the fuzzy approach can differ according to the severity of the data deviations. In particular, when the deviation is within the tolerance defined by the membership function, then the move-in-data has a smaller cost than the move-in-log: hence, there exists only one optimal alignment for these cases. For example, from Table \ref{t2_alignment} we can derive that for the first activity in 567 traces (744-177) the data deviation was indeed within the range, and hence we obtained only the move-in-data in output.

\begin{figure}[h]
\centering
\begin{minipage}[t]{0.48\textwidth}
\centering
\includegraphics[scale=0.6]{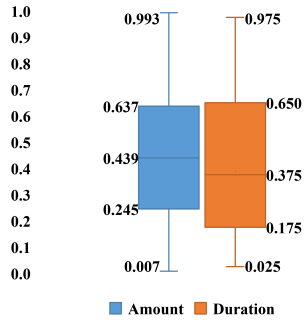}
\caption{Boxplots of severity of data deviation on both the data constraints}\label{devaition}
\end{minipage}
\begin{minipage}[t]{0.48\textwidth}
\centering
\includegraphics[scale=0.6]{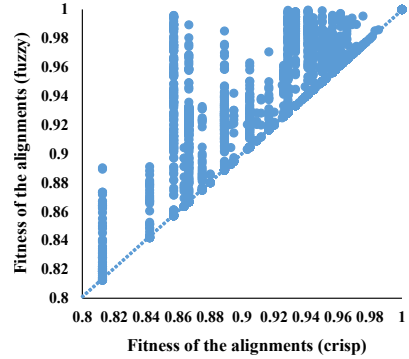}
\caption{Comparison of the fitness values obtained with crisp and fuzzy cost.}\label{fitness}
\end{minipage}
\end{figure}

Boxplots in Fig.\ref{devaition} show  the distributions of data deviation severity.We can see that the ranges are similar for both the constraints, with most of the values remaining below 0.65.
These distributions suggest that data deviations are mostly quite limited in our dataset; therefore, we expect relevant differences in fitness values computed by the fuzzy and the crisp approaches. 
Fig.~\ref{fitness} shows a scatter plot in which each point represents one trace. The x-axis is the fitness level of alignment with crisp costs, while the y-axis represents the value corresponding to the fuzzy cost. For all traces that are above the main diagonal,which amounts to 24.3\% of all traces, the fuzzy approach obtained higher values of fitness. For all traces on the main diagonal, the fitness level remains unchanged. 
\begin{table}
\centering
\caption{Number of different moves kinds for activities $W\_F\_C$ and $W\_FURTHER\_A$}\label{t2_alignment}
       \begin{tabular}{|l|l|l|l|l|}
	\hline
	 &\multicolumn{2}{|l}{$W\_F\_C\ with\ deviation$} & \multicolumn{2}{|l|}{$W\_FURTHER\_A\ with\ deviation$}\\
\hline
 & $\#$move-in-log &  $\#$move-in-data &  $\#$move-in-log &  $\#$move-in-data\\
Crisp  & 744 &  744 & 958 &  958\\
Fuzzy &  177 & 744 & 245 & 958  \\
\hline
	\end{tabular}
\end{table}
\begin{figure}
    \centering
    \includegraphics[scale=0.7]{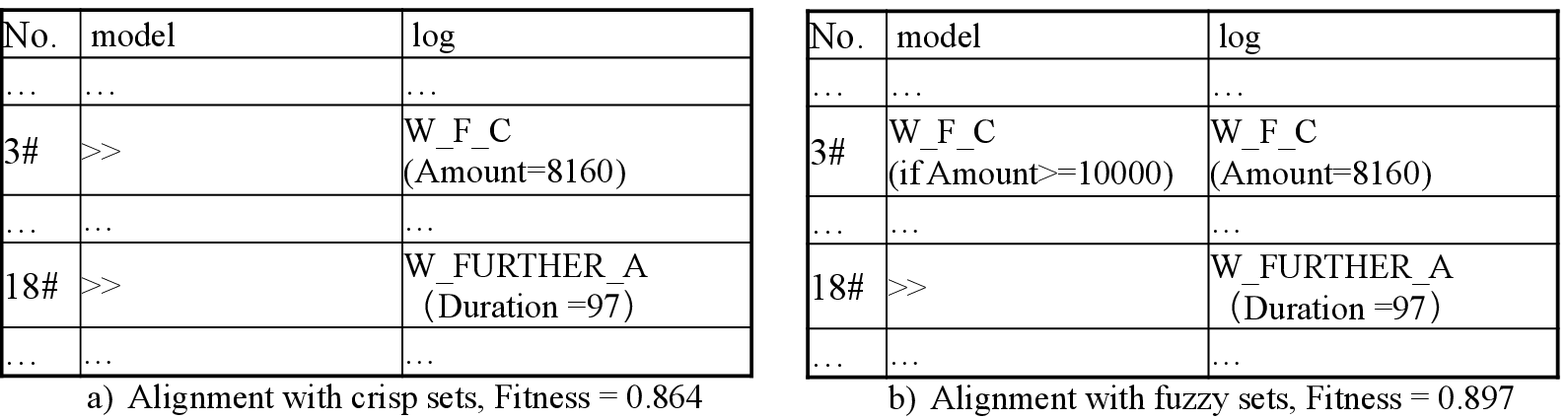}
    \caption{Differences between crisp and fuzzy alignment for $\sigma$.}
    \label{fig:exampleResult}
\end{figure}
In the following, we provide a practical example of the impact of the fuzzy cost on the diagnostic of a single trace.
\begin{example}
Let us consider $\sigma=\langle (A\_S,\{Amount$ $=8160\}),$ $(W\_FIRST\_A,\bot),$ $(W\_F\_C,\bot),$ $(A\_D,\bot),$ $(A\_A,\bot),$ $(A\_F,\bot),$ $(O\_S,\bot),$ $(O\_C,\bot),$ $(O\_S,\bot),$ \\ $(W\_C,\bot),$ $(O\_C,\bot),$ $(O\_S,\bot),$ $(W\_C,\bot),$ $(O\_C,\bot),$ $(O\_S,\bot),$ $(W\_C,\bot),$ \\ $(A\_R,\{Duration=97\}),$ $(W\_FURTHER\_A,\bot)$ $,(A\_{AP},\bot),$  $\rangle$. Fig.\ref{fig:exampleResult} shows the different alignments obtained adopting a crisp (Fig.\ref{fig:exampleResult}.a) and a fuzzy (Fig.\ref{fig:exampleResult}.b) cost function.  For the sake of space, here we report only the lines of the alignments related to the activities ruled by the data guards.For each move, we report the position of the move in the alignment followed by $"\#"$. Note that here we report the default optimal alignment returned by the tool, even though alternative alignments were possible. In particular, while for the second deviation multiple interpretations were returned by both the approaches, either as move-in-log or a move-in-data, since the amount of deviation is outside the tolerance range, the first deviations is always considered as a move-in-data in the fuzzy approach. Furthermore, the fuzzy approach returned a higher fitenss value for the trace than the crisp one; this is reasonable, since the first deviation is still close enough to the ideal value. 

\end{example}
Summing-up, the performed comparison did highlight how the use of a fuzzy cost led to improved diagnostics. In particular, the results show that the fuzzy approach allows to obtain a more fine-grained evaluation of traces compliance levels, allowing the analyst to differentiate between reasonably small and potentially critical deviations; furthermore, it allows to establish a preferred interpretation in cases in which the crisp function would consider possible options as equivalent, thus reducing ambiguities in interpretation.

	\section{Conclusion and Future work}
	\label{sec:conclusion}
	The present work investigated the use of fuzzy sets concepts in multi-perspective conformance checking. In particular, we shown how fuzzy sets notions can be used to take into account the severity of deviations when building the optimal alignment. We implemented the approach and performed a proof-of-concept over a synthetic dataset, comparing results obtained adopting a standard crisp logic and our fuzzy logic. The obtained results confirmed the capability of the approach of generating more accurate diagnostics, as shown both by a)the difference in terms of fitness of the overall set of executions, due to a more fine-grained evaluation of the magnitude of the occurred deviations, and b) by the differences obtained in terms of the different preferred explanations provided by the alignments of the different approaches.

Since this is an exploratory work, there are several research directions still to be explored. First, in future work we plan to test our approach in real-world experiments, to generalize the results obtained so far. Furthermore, as we mentioned in Section \ref{sec:methodology}, in the present work we investigated a fuzzy modeling only on the data perspective.
We plan to investigate this extension in future work. Similarly, we intend to address issues related to possible relations among data variables, incorporating this information to enhance the accuracy of the alignment. Another research direction we intend to explore, consists in introducing different \textit{aggregation} function; while here we used the classic sum operator to assess the overall trace conformance, in literature several fuzzy aggregation functions have been defined for membership functions, which can be used to tailor the cost function to the process analysts' needs. Finally, in future work we intend to investigate how to exploit our flexible conformance checking approach to enhance the system on-line resilience to exceptions and unforeseen events.

	\subsubsection{Acknowledgements.}
	The research leading to these results has received funding from the Brain Bridge Project sponsored by Philips Research.
	\bibliographystyle{splncs04}
	\bibliography{mybibliography}
\end{document}